\documentclass[10pt,twocolumn,letterpaper]{article}

\usepackage{wacv}              

\usepackage{graphicx}
\usepackage{amsmath}
\usepackage{amssymb}
\usepackage{booktabs}
\usepackage{adjustbox}
\usepackage{pifont}
\newcommand{\cmark}{\ding{51}}
\newcommand{\xmark}{\ding{55}}
\usepackage{multirow}
\usepackage{float}

\usepackage[pagebackref,breaklinks,colorlinks]{hyperref}

\usepackage[capitalize]{cleveref}
\crefname{section}{Sec.}{Secs.}
\Crefname{section}{Section}{Sections}
\Crefname{table}{Table}{Tables}
\crefname{table}{Tab.}{Tabs.}

\def\wacvPaperID{2728} 
\def\confName{WACV}
\def\confYear{2025}

\begin{document}

\renewcommand{\arraystretch}{0.9}  

\title{DDS: Decoupled Dynamic Scene-Graph Generation Network}

\author{
    A S M~Iftekhar\textsuperscript{*}\textsuperscript{\ddag}, 
    Raphael~Ruschel\thanks{\parbox[t]{\textwidth}{represents equal contribution \\ Iftekhar is currently with the Responsible AI team, Microsoft}} \textsuperscript{\ddag}, 
    Satish Kumar\textsuperscript{\ddag}, 
    Suya You\textsuperscript{\S}, 
    B. S. Manjunath\textsuperscript{\ddag} \\
    \textsuperscript{\ddag} University of California Santa Barbara \\ 
    \textsuperscript{\S} Army Research Laboratory, Intelligent Perception, CISD, Los Angeles, CA\\ 
    \tt \small (iftekhar, raphael251, satishkumar, manj)@ucsb.edu, \\
    \tt \small suya.you.civ@army.mil
}

\maketitle

\newif\ifdrafting
\draftingfalse 
\ifdrafting
\newcommand{\ER} [1] {\textcolor{red}{#1}}
\else
\newcommand{\ER} [1] {\textcolor{black}{#1}}
\fi


\begin{abstract}
Scene-graph generation involves creating a structural representation of the relationships between objects in a scene by predicting subject-object-relation triplets from input data. Existing methods show poor performance in detecting triplets outside of a predefined set, primarily due to their reliance on dependent feature learning. To address this issue we propose DDS-- a decoupled dynamic scene-graph generation network-- that consists of two independent branches that can disentangle extracted features. The key innovation of the current paper is the decoupling of the features representing the relationships from those of the objects, which enables the detection of novel object-relationship combinations. The DDS model is evaluated on three datasets and outperforms previous methods by a significant margin, especially in detecting previously unseen triplets. 

\end{abstract}
\section{Introduction}
Dynamic Scene-Graph (DSG) provides a graph structure presenting the relationships among different objects in a scene by predicting relationship triplets composed of $\langle subject, object, relationship \rangle$ at each frame of an input video. This acts as a foundational block for various computer vision tasks\cite{ulutan2020actor, liu2019caesar}. Current DSG generation systems~\cite{ji2021detecting, li2022dynamic, cong2021spatial} have poor performance when facing a triplet that was not present during training, even though the individual components have been seen. However, in a more realistic deployment scenario, the network will likely encounter triplets that it has not seen before. Therefore, a system should be able to transfer the learned concepts of relationships and objects to compose unseen triplets. This poor performance is mainly attributed to the learning of highly dependent feature representations of relationships and objects. The proposed decoupled dynamic scene-graph (DDS) addresses this issue. 

\begin{figure}[t]
\begin{center}
\includegraphics[width=\linewidth]{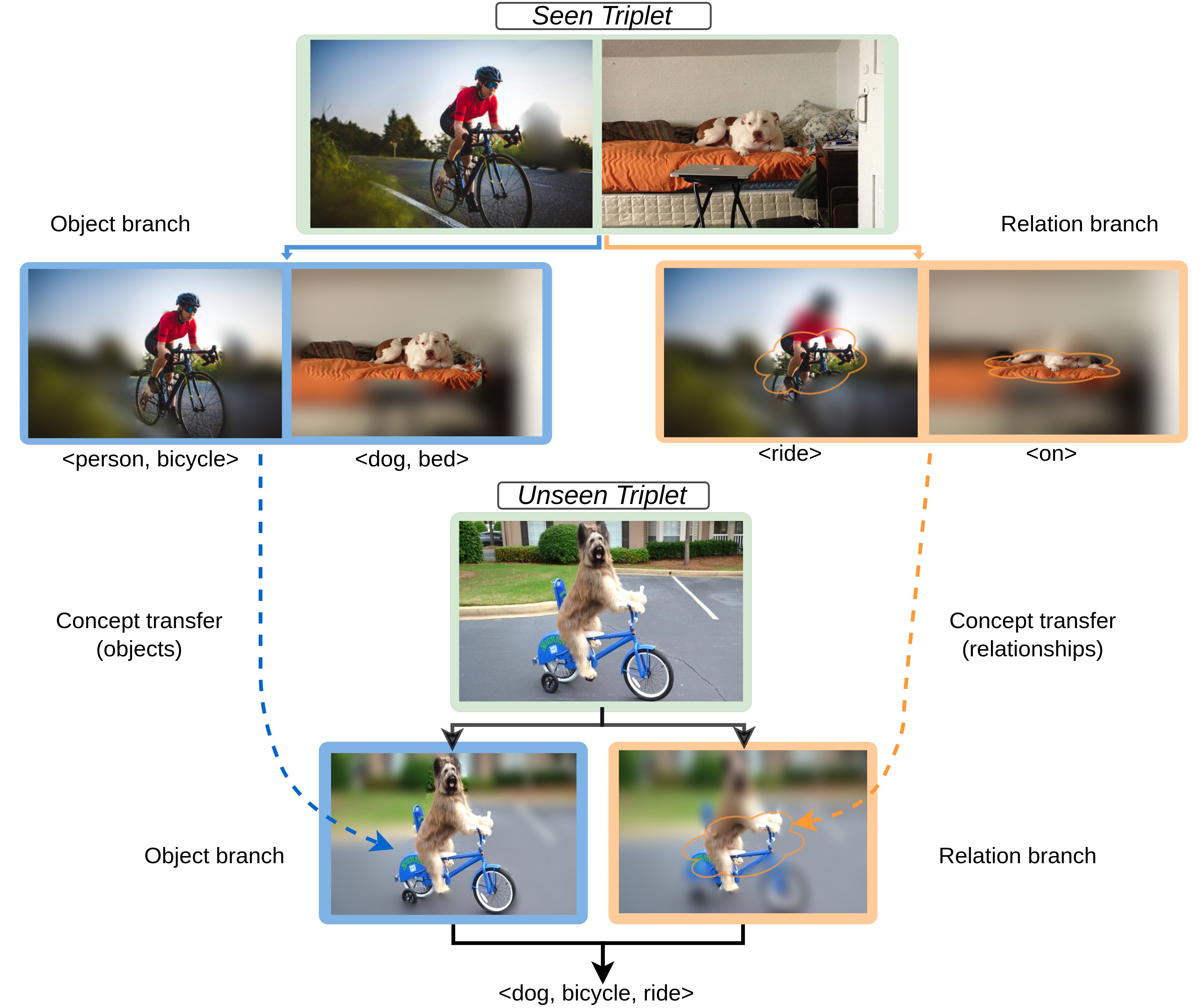}
\end{center}
\vspace{-0.4cm}
\caption{Diagram to show the concept learning and transferring in DDS. By focusing on different spatial regions, DDS learns the concept of relationships (ride, on) and objects (person, bicycle, bed) independently.
}
\label{fig:concept}
\vspace{-0.2cm}
\end{figure}

Fig. \ref{fig:concept} shows the core idea of the proposed DDS network. This architecture utilizes two different branches to learn decoupled features for relationships and objects. As shown in the figure, DDS learns the concept of `ride', `on', `person', `bicycle', and `bed' from the training examples of a `person riding a bike' and a `dog on the bed' independently. The decoupled design makes DDS look into different spatial regions for relationships and objects. These learned concepts are transferred to successfully detect the unseen triplet $\langle dog, bicycle, ride \rangle$.

DDS ensures the learning of discriminative spatio-temporal cues for relationships and objects. Fig. \ref{fig:model_architecture} shows the overview of our architecture. It consists of two separate branches: the relation and the object branch. We chose a transformer-based encoder-decoder \cite{carion2020end} architecture for these branches with two different sets of queries. Moreover, a novel temporal decoder is added to embed temporal information into the queries. These separate sets of queries focus on learning generalized representations for relationships and objects from differently encoded feature maps in both temporal and spatial domains. This is significantly better than the existing works, where the same object features are used for both object and relationship detection. Also, DDS does not have the dependence on off-the-shelf object detectors like previous works.
 
 Our proposed model is thoroughly evaluated on the Action-Genome\cite{ji2020action} dataset for DSG generation, where it achieves significant performance gains compared to the SOTA models. Additionally, we evaluate DDS on the task of static scene-graph (SSG) generation on the HICO-DET\cite{chao2018learning} dataset and unusual SSG generation on the UnRel\cite{peyre2017weakly} dataset, where DDS outperforms all the existing models in both datasets. Finally, the proposed design choices are evaluated in an extensive ablation study. 

\section{Related Works}
DDS is built on the previously developed works in SSG and DSG generation. This section is used to review the literature in the mentioned areas along with additional relevant publications on scene-graph generation under the compositional setting.

\ER{\subsection{Static Scene-Graph (SSG) Generation}}
SSG generation is proposed for the task of image retrieval. The initial works rely heavily on two-stage (object detection and then scene-graph generation) structures. Also, many authors utilize prior knowledge\cite{ulutan2020vsgnet, liu2020consnet, iftekhar2021gtnet} (e.g. semantic knowledge, statistical heuristics) for SSG generation. Despite recent improvements in SSG generation, these methods are heavily constrained by their reliance on the object detection quality as noted in\cite{iftekhar2022look}. 
 
 Modern works in SSG generation focus on utilizing a one-stage Transformer based architecture to deal with the aforementioned issues. These architectures rely on set-based predictions to generate SSG. Among these works, Qpic\cite{tamura2021qpic} uses a single encoder-decoder model while CDN\cite{zhang2021mining} extends Qpic by using sequential decoding of objects and relationships. Additionally, MSTR\cite{kim2022mstr} enables the use of multi-scale feature maps to these networks. Another concurrent work,  SSRT\cite{iftekhar2022look} refines the overall architecture with spatial and semantic support features. Moreover, a recent line of research heavily exploits the usage of very large-scale semantic knowledge engines (e.g. CLIP\cite{radford2021learning})\cite{qu2022distillation, iftekhar2022look, liao2022gen}. Apart from the obvious limitation of these works being unable to utilize temporal dependencies, they perform poorly while detecting unseen triplets. With the decoupled multi-branch design, we significantly differ from these works by using separate sets of queries for relationship and object detection. 
 
\ER{\subsection{Dynamic Scene-Graph (DSG) Generation}}
 DSG is an extension of the SSG where the scene-graph is created for videos. This process is harder since temporal cues need to be utilized\cite{ji2021detecting, li2022dynamic, cong2021spatial}. Current works in this area have two-stage architectures following the initial works on SSG. Among these works, STTran\cite{cong2021spatial} utilizes a temporal decoder based on the self-attention mechanism. DSGAP\cite{li2022dynamic} expands STTran with an anticipatory pre-training paradigm. On the other hand, HORT\cite{ji2021detecting} utilizes a multi-branch design with different types of Transformers. 
Both STTran and HORT use similar features for relationship and object detection. These features come from the object bounding boxes predicted by off-the-shelf object detectors. However, using similar features for relationship and object detection forces the learning of relationships and objects to be dependent on each other.

\subsection{Compositionality in Scene-Graph Generation}

Creating new compositions from base known concepts during inference is known as compositional zero-shot learning (CZSL) under the compositional setting\cite{naeem2021learning, kumar2022locl}. In this paper, we utilize this setting to evaluate our model. Kato et al. \cite{kato2018compositional} introduce CZSL in SSG generation with an embedding-based model. Many following works\cite{hou2021atl, hou2022scl, hou2021detecting} adapt different object-affordance ideas. These works assume there exist common relationships between the subjects and the objects. This work does not have such limited assumptions, and as a result, can generate scene graphs even when the relationships are very unusual (See Table \ref{tab:map_unrel}). 

\begin{figure*}[t]
\begin{center}
\includegraphics[width=\linewidth]{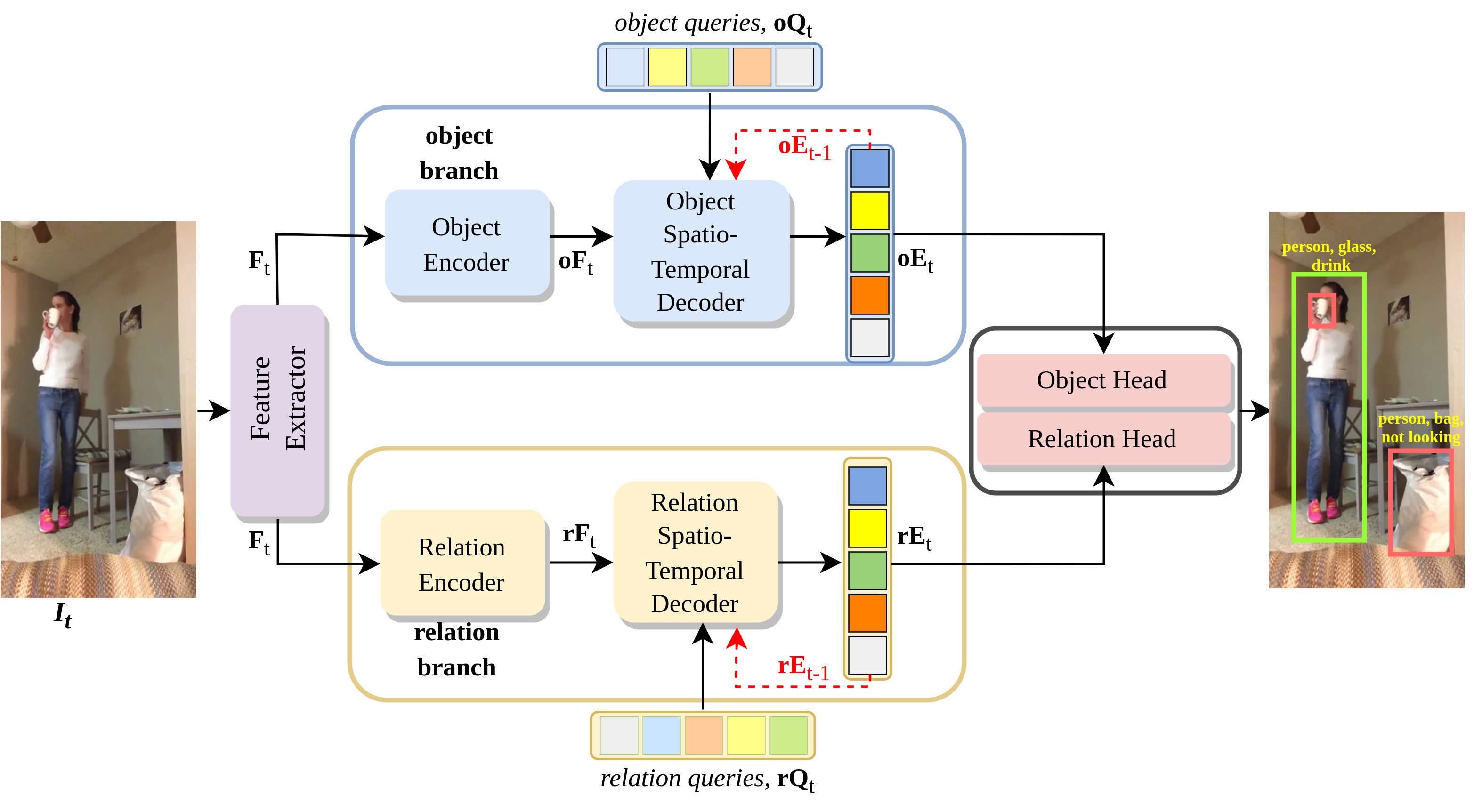}
\end{center}
\vspace{-0.4cm}
\caption{Overview of DDS's architecture. Given an input frame $\mathbf{I}_{t}$, features are extracted by the backbone and fed to decoupled object and relation branches, each with an encoder and spatio-temporal decoder. The decoders process queries and previous frame embeddings (red arrow) to produce learned embeddings, which are used by the object and relation heads to predict relationship triplets. 
}
\label{fig:model_architecture}
\vspace{-0.2cm}
\end{figure*}

\section{Method} 
We propose a multi-branch network to address the challenge of dependent feature learning in DSG generation, which decreases the performance of detection of unseen relationship triplets in current models. The network learns distinct feature representations for relationships and objects. Before detailing the architecture, we first formulate the problem.

\subsection{Problem Formulation}
Given an input video, $\mathbf{V}=\{\mathbf{I}_{1}, \mathbf{I}_{2},.., \mathbf{I}_{t},.., \mathbf{I}_{T}\}$ with $T$ frames, the task in DSG generation is to predict a set of relationship triplets, $\{R_{1}, R_{2},..,R_{t},..,R_{M}\}$ at every frame of the video. Every frame has $N_{M}$ number of relationship triplets. Each relationship triplet can be presented by $\langle s, o, r_{so}\rangle$. Here, $s$, $o$ refers to subject, object and are represented by bounding boxes and category labels. $r_{so}$ is the relationship between $s$ and $o$. In a single frame $\mathbf{I}_{t}$, $s$ and $o$ can have multiple relations, as shown in the sample input-output pair in Figure \ref{fig:model_architecture}.  

In this paper, the main goal is to predict relationship triplets under the compositional setting. In this setting, the networks see all subjects, objects, and relationships during training but not all combinations available, leaving a portion to be exclusive only to the test set so we can evaluate the performance on unseen triplets. 

\subsection{Technical Overview}
The proposed work adopts a one-stage approach for DSG generation compared to the current two-stage methods\cite{ji2021detecting, li2022dynamic, cong2021spatial} as the former \cite{kim2021hotr,tamura2021qpic, chen2021reformulating,zou2021end} have shown impressive performance in creating SSG. However, these image-based works present poor generalization capabilities. Therefore, we propose a network that uses a different set of queries with two branches, where each branch follows a Transformer-like encoder-decoder architecture, similar to \cite{kumar2023methanemapper, kumar2024wildlifemapper}. Fig. \ref{fig:model_architecture} shows a diagram of the model, where a convolutional neural network (CNN) extracts features from the input frame, and those are encoded differently by the object and the relation encoders. Each spatio-temporal decoder takes encoded features from their respective encoder in addition to two types of inputs: queries for the current frame ($\mathbf{oQ}_{t}$, $\mathbf{rQ}_{t}$) and the embeddings ($\mathbf{oE}_{t-1}$, $\mathbf{rE}_{t-1}$) propagated from the previous frame. As the encoded features differ for each branch, the queries learn decoupled features for relationships and objects. The decoder outputs are the learned object and relation spatio-temporal embeddings and are used as input to the relation and the object heads for final predictions.       

\subsection{Feature Extraction \& Encoders}\label{sec:feat_enc}
Consider a frame $\mathbf{I}_{t}\in \mathbb{R}^{N_{C} \times H \times W}$ at time t of the input video $V$. A CNN network is used as backbone, $\mathbf {B}$ (e.g. resnet-50\cite{he2016deep}) to extract features that are then reduced using $1 \times 1$ convolutions, flattened, and added positional embeddings to get the feature map $\mathbf {F}_{t} \in \mathbb{R}^{(H'W') \times d'}$ that is used as a common feature for both the relation and the object branches.  

Both of the network's branches have an encoder comprising of stacked multi-head self-attention layers\cite{vaswani2017attention} with a feed-forward network (FFN). The output of the encoders are two separate feature maps, $\mathbf{rF}_{t}$, and $\mathbf{oF}_{t}$, which represent the features from the relationship and object branch, respectively.

\subsection{Input Queries}\label{sec:queries}
The DDS framework consists of 2 independent branches with each one containing one decoder. Each decoder takes a set of randomly initialized, trainable queries ${\mathbf{rQ}_{t}}$ and ${\mathbf{oQ}_{t}}$ for the relationship and object branch, respectively, and process them through a dual-stage process. This process, designed to explore spatio-temporal features, begins with the temporal decoders aggregating information across frames, followed by spatial decoders enriching this information with the current frame context.

\begin{figure*}[t]
\begin{center}
\includegraphics[width=0.9\linewidth]{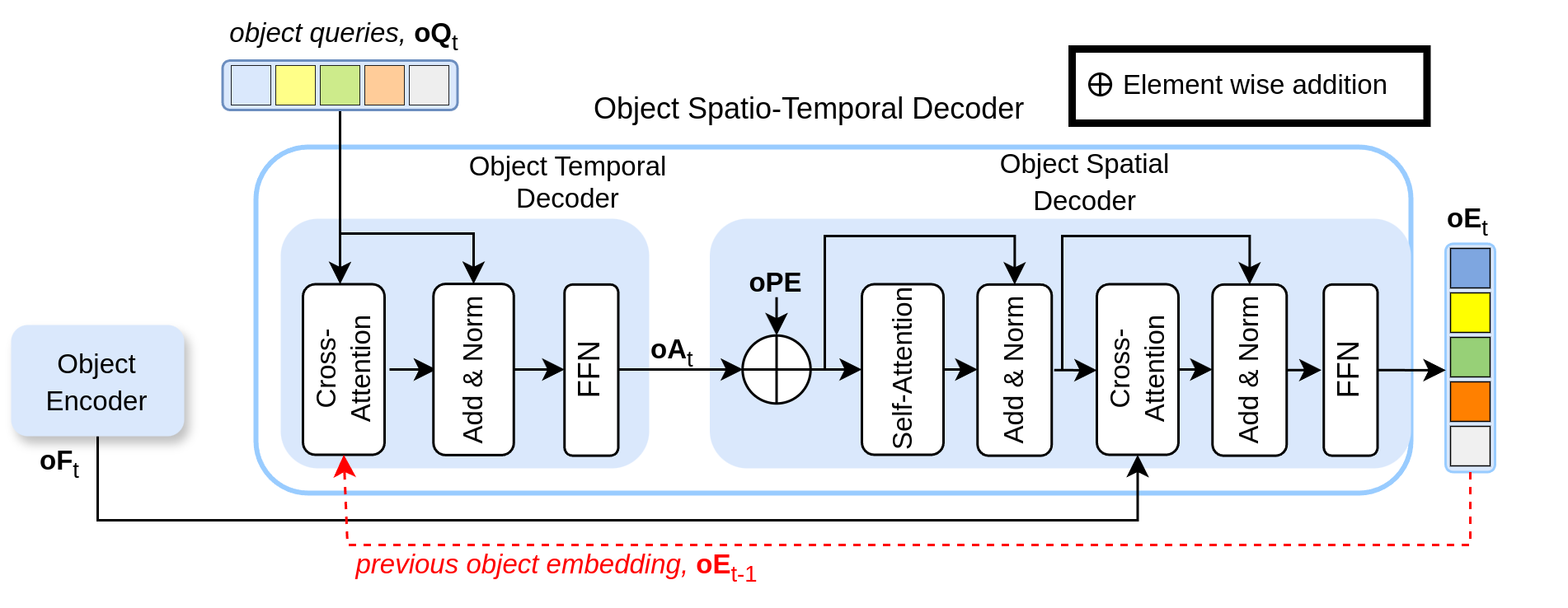}
\end{center}
\vspace{-0.4cm}
\caption{Design of the spatio-temporal decoders using the Object decoder as an example. The relationship decoder uses the same architecture, however, with its corresponding inputs adjusted.      
}
\label{fig:detail_architecture}
\vspace{-0.4cm}
\end{figure*} 

\subsection{Spatio-Temporal Decoders}\label{sec:spatemp_decoder}
After encoding the features using a standard transformer encoder, DDS uses a two-stage decoding process, where each decoder consists of two small components: temporal and spatial decoder, which are shown in Fig. \ref{fig:detail_architecture} for reference. The proposed multi-branch design ensures discriminative feature learning for the queries of each branch.    

    
\textbf{Temporal Decoders:} These decoders allow queries to leverage temporal dependencies. 
Each temporal decoder takes two sets as inputs: the current frame's queries and the embeddings from the previous frames. For frame $\mathbf{I}_t$,  the current frame's relation and object queries sets are defined as ${\mathbf{rQ}_{t}} \in \mathbb{R}^{N_{q} \times d}$  and $ \mathbf{oQ}_{t} \in \mathbb{R}^{N_{q} \times d}$. Every query is a $d$ dimensional vector, and every branch has $N_{q}$ number of queries. Embeddings from the previous frames for the relation and the object branches are presented by ${ \mathbf{rE}_{t-1}} \in \mathbb{R}^{N_{q} \times d}$  and ${ \mathbf{oE}_{t-1}} \in \mathbb{R}^{N_{q} \times d}$, and are marked with a red arrow in Fig.\ref{fig:detail_architecture}. 

The cross-attention in the temporal decoders allows the current frame's queries to select what to learn from the previous frame's embeddings. The outputs of the temporal decoders are the temporally aggregated queries $ \mathbf{rA}_{t}$, $\mathbf{oA}_{t}$. They are fed to their respective spatial decoders. In the case of the first frame of a video, the temporal decoders directly output $\mathbf{rQ}_{t}$ and $\mathbf{oQ}_{t}$ as $\mathbf{rA}_{t}$ and $\mathbf{oA}_{t}$ without passing them through the cross-attention and FFN blocks as there is no previous frame. 

\textbf{Spatial Decoders:} The spatial decoders architecture is similar to the standard Transformer decoder\cite{carion2020end}. These decoders consist of both self-attention and cross-attention layers along with FFN networks. Each decoder takes encoded feature maps ($\mathbf{rF}_{t}$ or $\mathbf{oF}_{t}$) along with the aggregated queries of the temporal decoders ($\mathbf{rA}_{t}$ or $\mathbf{oA}_{t}$) from their respective branch as inputs. Also, these decoders take learnable positional embeddings, $\mathbf{rPE}$, and $\mathbf{oPE}$, for the relationship and object decoder, respectively. These embeddings for the relation and the object branch are defined as $\mathbf{rPE} \in \mathbb{R}^{N_{q} \times d} $, $\mathbf{oPE} \in \mathbb{R}^{N_{q} \times d}$.  
\begin{flalign}
    &\mathbf{rE}_{t} = \text{Relation Decoder}(\mathbf{rN}_{t}, \mathbf{rPE},\mathbf{rQ}_{t}) \\
    &\mathbf{oE}_{t} = \text{Object Decoder}(\mathbf{oN}_{t},  \mathbf{oPE}, \mathbf{oQ}_{t})
\end{flalign}
The outputs of the decoders are the learned spatio-temporal embeddings. They are used in the object and the relation heads to make the final relationship triplet predictions.

\subsection{Object \& Relation Heads} \label{sec:obj_head}
The output embeddings from the object spatio-temporal decoder, $\mathbf{oE}_{t}$ are fed to four different FFNs. For input frame $\mathbf{I}_{t}$, these FFNs predict subject bounding boxes, $\mathbf{sB}_{t} \in [0,1]^{N_{q} \times 4}$, object bounding boxes, $\mathbf{oB}_{t} \in [0,1]^{N_{q} \times 4}$, subject prediction vectors, $\mathbf{sP}_{t} \in [0,1]^{N_{q} \times O}$, and object prediction vectors, $\mathbf{oP}_{t} \in [0,1]^{N_{q} \times N_{o}}$. Here, $N_{q}$ is the number of queries, and $N_{o}$ is the total number of objects.
Similarly, $\mathbf{rE}_{t}$, are fed to two FFNs that produce as output the relation prediction vectors, $\mathbf{rP}_{t} [0,1]^{N_{q} \times N_{r}}$ and relation region bounding boxes $\mathbf{rB}_{t} \in [0,1]^{N_{q} \times 4}$. Notice that the relation region bounding box is defined as the union between the subject and object bounding boxes, and is used solely during training. 

\subsection{Inference} 
During inference, we take the outputs from the object and relation heads as in \ref{sec:obj_head} and compose $N_{q}$ relationship pairs by leveraging the maximum confidence score in $\mathbf{sP}_{t}$ and $\mathbf{oP}_{t}$. The maximum confidence score is used to create $\mathbf{sP}_{tmax} \in [0,1]^{N_{q}} $ and $\mathbf{oP}_{tmax} \in [0,1]^{N_{q}}$ and the corresponding index is used to determine the category label for each of the bounding boxes. For every composed relationship pair, the final relation score is calculated as a multiplication of each category's confidence score



\subsection{Training} \label{sec:training}
For training DDS, we utilize losses similar to Qpic\cite{tamura2021qpic}. This loss calculation implicitly binds the two sets of queries from the relation and the object branch. The loss calculation happens in two stages:

In the first stage, we find the bipartite matching between the predictions $\mathbf{P}_{t}$ for frame $\mathbf{I}_{t}$ and the ground truths $\mathbf{G}_{t}$. One important detail to note here is that there are three kinds of ground truth bounding boxes: subject bounding boxes, object bounding boxes, and relation regions. Ground truth relation regions refer to the union bounding boxes between the subject and the object bounding boxes for the current interaction and are only used during the training phase.
The matching cost metric is defined as: 
\begin{equation}
 {\mathbf{C}^{(i,j)}} = \eta_b(\mathbf{C}^{(i,j)}_{sb}+\mathbf{C}^{(i,j)}_{ob}+ \mathbf{C}^{(i,j)}_{rb}) +\eta_o\mathbf{C}^{(i,j)}_{o}+\eta_r\mathbf{C}^{(i,j)}_{r}
 \label{eq: hung_loss}
\end{equation}
Where $\mathbf{C}^{(i,j)}_{sb}, \mathbf{C}^{(i,j)}_{ob}, \mathbf{C}^{(i,j)}_{rb} $ are the subject bounding box, the object bounding box and the relation region matching costs, $\mathbf{C}^{(i,j)}_{o}$ is the object label matching cost, and  $\mathbf{C}^{(i,j)}_{r}$ is the relation label matching cost between $i^{th}$ element from $\mathbf{P}_{t}$ and $j^{th}$ element from $\mathbf{G}_{t}$. These costs are calculated following\cite{tamura2021qpic}. $\eta_b, \eta_o, \eta_r$ are fixed hyper-parameters. The Hungarian matching algorithm\cite{carion2020end} is used to find the optimal matching between the predictions and the ground truths by using these cost metrics.   
After this matching, every prediction is associated with a ground truth. Next, the following loss is calculated for training the network:
\begin{equation}
 \mathcal{L} =\lambda_g\mathcal{L}_{GIOU} +\lambda_l\mathcal{L}_{L1}
 +\lambda_o\mathcal{L}_{obj}
 +\lambda_r\mathcal{L}_{rel},
 \label{eq: overll_loss}
\end{equation}
Here, $\mathcal{L}_{GIOU}$ and $\mathcal{L}_{L1}$ are the generalized intersection over union (gIOU) and L$1$ box regression losses for the predicted subject bounding boxes, object bounding boxes, and relation regions. $\mathcal{L}_{obj}$ is the cross-entropy loss for subject and object label predictions. $\mathcal{L}_{rel}$ is the binary cross-entropy loss for the relationship label predictions. $\lambda_o$, $\lambda_g$, $\lambda_l$, and $\lambda_r$ are the corresponding hyper-parameters. 

Notice that a portion of the datasets \cite{ji2020action, chao2018learning} fixes the subject as humans. In this case, the subject prediction vectors and subject bounding boxes are not used for loss calculation. 

\section{Experiments}

\subsection{Experimental Setup} \label{sec:exp_setup}
To evaluate DDS's performance on video data, we use the Action Genome (AG)\cite{ji2020action} dataset. Moreover, we show our model's performance in SSG generation datasets: HICO-DET\cite{chao2018learning} and UnRel\cite{peyre2017weakly}. \ER{As these datasets only contain images, each sample is treated as a single-frame video, and thus the only difference is that the feedback mechanism will not be used.} Next, the used datasets will be presented in more detail:  



\textbf{Action Genome (AG)\cite{ji2020action}:} This dataset is built on top of the Charades\cite{sigurdsson2016hollywood} dataset, provides frame-level annotations, and is extensively used in the literature for DSG generation. It has 36 distinct object classes and 25 relationship classes. The object classes are common household items such as doors, windows, and cups, and have a total (train and test set) of $476,229$ bounding boxes. In total, AG provides $1,715,568$ instances of the mentioned classes contained in $135,484$ subject-object pairs. Every subject-object pair can have multiple relations. Also, on AG the subject class is always human.

Originally, the AG dataset provided $7,464$ videos with $166,785$ frames in the training set and $1,737$ videos with $54,371$ frames in the test set. The original training set contains $530$ relationship triplets. All these relationship triplets are present in the test set. We refer to this setting as the fully-supervised setting. As the main interest in this paper is to evaluate DDS's performance in the compositional setting, a new training split of the data is created. This new proposed training set contains $6,784$ videos with $146,517$ frames containing $421$ relationship triplets. The original test set is not changed. It contains $499$ object-relationship, where $80$ of them are not present in our new training set. 

\textbf{HICO-Det\cite{chao2018learning}:} This dataset has $80$ objects and $117$ relationship classes. In the literature, this dataset is used for evaluating SSG generation performance under compositional setting\cite{hou2021detecting, hou2021atl, hou2022scl, hou2021vcl}. DDS's performance is reported following the RF (Rare First) protocol provided by \cite{hou2021vcl}. This protocol has $37,328$ images in the training set with $480$ relationship triplets. The test set has $9,552$ images with $480$ seen relationship triplets and $120$ unseen relationship triplets.  

\textbf{UnRel\cite{peyre2017weakly}:} This dataset provides extremely unusual SSG triplets, for example: ($\langle elephant, bike, riding \rangle$). It has $4000$ training and $1000$ test images with $100$ objects and $70$ relationships. The original train/test split provided by the authors already provides a compositional setting where the test set has $65$ unseen relationship triplets. The training set contains $4000$ seen relationship triplets.


\subsection{Evaluation Metrics}
Following existing works \cite{ji2020action, ji2021detecting, li2022dynamic, cong2021spatial}, we report our performances in AG dataset with Recall@K metric with $K=[20,50]$. \ER{We utilize only the SGDet\cite{ji2020action} protocol to report our performances, as PREDCLS and SGPRED are not applicable to one-stage methods. In this protocol, the network needs to detect relationship triplets along with subject and object bounding boxes. We carry out experiments on two different scenarios: \textit{With constraint}: Each query is only allowed to predict a single triplet, and \textit{No constraint}: Where each \textit{subject-object} pair can generate multiple verb predictions. Given that AG presents a long-tailed distribution over the relationship classes, recall alone might result in biased results if the model is favorable towards the most frequent classes. To provide a more complete analysis, we also present the mean Recall@k (mR@K), where the recall of each predicate is measured and then averaged over the number of classes. Moreover, mAP (mean average precision) was selected to report the performances in UnRel and HICO-Det datasets similar to current works\cite{peyre2017weakly, peyre2019detecting, hou2021atl, hou2021detecting}}. Here, performances are reported in three categories: unseen (only unseen relationship triplets), seen (only seen relationship triplets), and full (all relationship triplets)~\cite{hou2021vcl}. For all datasets, a prediction triplet from DDS is considered correct if subject and object bounding boxes have at least $0.5$ Intersection over Union (IoU) with ground truth bounding boxes, and subject, object, and relationship labels match with ground truth labels.

\subsection{Implementation details}  
ResNet-50\cite{he2016deep} is used as the CNN backbone. Both temporal decoders inside the spatio-temporal decoders have a single layer. We follow  Qpic's \cite{tamura2021qpic} setup for the encoders.  We select $6$ layers for the object spatial decoder with $3$ layers for the relation spatial decoder. All loss coefficients in equation \ref{eq: hung_loss} and equation \ref{eq: overll_loss} are set as \cite{tamura2021qpic}.  The number of queries in each branch is $64$. Each query is a vector of size $256$. The model is trained with AdamW\cite{ilya2019decoupled} optimizer and uses the training strategy in \cite{ruschel2023bload} for efficiently handling video sequences. We initialize the parameters of DDS from DETR\cite{carion2020end} trained on COCO\cite{lin2014microsoft} object detection dataset. The initial learning rate for the backbone network is $10^{-6}$ and for the other part of the network is $10^{-5}$. All evaluations are done following the codebase provided by \cite{tang2019learning}. 

\begin{table*}[ht]
\centering
\small
\setlength{\tabcolsep}{4pt}
\begin{tabular}{@{}lcccccccccccc@{}}
\toprule
& \multicolumn{12}{c}{SGDet} \\
\cmidrule(lr){2-13}
& \multicolumn{6}{c}{\ER{With constraint}} & \multicolumn{6}{c}{No constraint} \\
\cmidrule(lr){2-7} \cmidrule(l){8-13}

Method & \ER{mR@10} & \ER{mR@20} & \ER{mR@50} & \ER{R@10} & \ER{R@20} & \ER{R@50}  & \ER{mR@10} & \ER{mR@20} & \ER{mR@50} & R@10 & R@20 & R@50 \\ 
\midrule
ReIDN \cite{zhang2019graphical} & \ER{3.3} & \ER{3.3} & \ER{3.3} & \ER{9.1} & \ER{9.1} & \ER{9.1}  & \ER{7.5} & \ER{18.8} & \ER{33.7} & 13.6 & 23 & 36.6  \\
TRACE \cite{teng2021target} & \ER{8.2} & \ER{8.2} & \ER{8.2} & \ER{13.9} & \ER{14.5} & \ER{14.5}  & \ER{22.8} & \ER{31.3} & \ER{41.8} & 26.5 & 35.6 & 45.3  \\
iSGG \cite{khandelwal2022iterative} & \ER{-} & \ER{19.7} & \ER{22.9} & \ER{-} & \ER{29.2} & \ER{35.3} & \ER{-}  & \ER{-} & \ER{-} & - & - & -  \\
STTran \cite{cong2021spatial} & \ER{16.6} & \ER{20.8} & \ER{22.2} & \ER{25.2} & \ER{29.1} & \ER{37}  & \ER{20.9} & \ER{29.7} & \ER{39.2} & 24.6 & 36.2 & 48.8  \\
\ER{STTran-TPI} \cite{STTran-TPI} & \ER{15.6} & \ER{20.2} & \ER{21.8} & \ER{26.2} & \ER{29.1} & \ER{34.6}  & \ER{-} & \ER{-} & \ER{-} & \ER{25.7} & \ER{37.9} & \ER{50.1} \\
\ER{APT} \cite{li2022dynamic} & \ER{-} & \ER{-} & \ER{-} & \ER{26.3} & \ER{29.1} & \ER{38.3}  & \ER{-} & \ER{-} & \ER{-} & \ER{-} & \ER{-} & \ER{-}  \\
\ER{TEMPURA} \cite{TEMPURA} & \ER{18.5} & \ER{22.6} & \ER{23.7} & \ER{28.1} & \ER{33.4} & \ER{34.9}  & \ER{24.7} & \ER{33.9} & \ER{43.7} & \ER{29.8} & \ER{38.1} & \ER{46.4}  \\
\ER{VsCGG} \cite{VsCGG} & \ER{18.7} & \ER{-} & \ER{24.2} & \ER{27.4} & \ER{35.8} & \ER{38.2}  & \ER{24.3} & \ER{33.1} & \ER{42.8} & \ER{29.3} & \ER{40.2} & \ER{48.9}  \\
\ER{TD2-Net (p)} \cite{Lin2024TD2NetTD} & \ER{20.4} & \ER{-} & \ER{23} & \ER{26.1} & \ER{28.7} & \ER{37.1}  & \ER{27.9} & \ER{33} & \ER{46.3} & \ER{30.5} & \ER{-} & \ER{49.3}  \\
DSG-DETR\cite{feng2023exploiting} & - & - & - & 30.3 & 34.8 & 36.1 & - & - & - & 32.1 & 40.9 & 48.3 \\
OED\cite{wang2024oed} & - & - & - & 33.5 & 40.9 & \textbf{48.9} & - & - & - & 35.3 & \textbf{44} & \textbf{51.8} \\
\textbf{DDS (ours)} & \textbf{24.5} & \textbf{29.1} & \textbf{32.2} & \textbf{36.2} & \textbf{42.0} & 47.3 &  \textbf{32.9} & \textbf{41.3} & \textbf{48.7} & \textbf{37.3} & 43.3 & 51.5 \\

\bottomrule
\end{tabular}
\vspace{0.2cm}
\caption{DDS's performance comparison in AG test set. Here, like other models, DDS is trained in the full training set of AG dataset. The best results are shown in \textbf{bold}. For the other models, all the reported numbers are taken from the original publications.}
\label{tab:map_AG}
\end{table*}

\begin{table}[t]
\centering
\adjustbox{width=0.8\columnwidth}{
\begin{tabular}{lcccc}
        & \multicolumn{2}{c}{Seen}        & \multicolumn{1}{r}{Unseen} & \multicolumn{1}{r}{} \\ \hline
        & R@20           & R@50           & R@20                       & R@50                 \\ \hline
STTran\cite{cong2021spatial}* & 33.7          & 36.6          & 0.3                     & 4.4                 \\
DDS (Ours)     & \textbf{41.8} & \textbf{48.8} & \textbf{7.4}              & \textbf{18.2}       \\ \hline
\end{tabular}
}
\vspace{0.2cm}
\caption{DDS's performance comparison in AG test set under the compositional setting. Both reported models are trained on the proposed small-size training set under the compositional setting.  * means the model was trained using publicly available code. Among recent DSG generation models, only STTran's\cite{cong2021spatial} code is publicly available. The best results are shown in \textbf{bold}.}
\vspace{-0.2cm}
\label{tab:ag_sota_comp}
\end{table}

When training in the AG dataset, we drop the learning rate by $10$ times at every $40$ epochs and utilize a batch size of $128$. DDS processes each video frame from a single video sequentially. We utilize scale augmentation like\cite{carion2020end}. Input frames are resized with the shortest side being at least 480 and at most 800 pixels, and longest side is at most 800. 

In the other datasets\cite{chao2018learning, peyre2017weakly}, the learning rate is dropped by $10$ times at every $60$ epochs with a batch size of $16$. We use a scale augmentation scheme similar to the one used for AG, except that the longest side of the resized image is chosen as $1333$. The training schedule is selected based on the convergence of losses. Upon acceptance, we will publicly release our trained models and code.

\section{Results \& Analysis}
 \begin{table}[t]
\centering
\adjustbox{width=\columnwidth}{

\begin{tabular}{lccc}
\hline
Method                                & Unseen (mAP) & Seen (mAP) & Full (mAP) \\ \hline
VCL\cite{hou2021vcl} & 10.1                           & 24.3                          & 21.4                          \\
ATL\cite{hou2021atl} & 9.2                             & 24.7                          & 21.6                          \\

FCL\cite{hou2021detecting} & 13.2                            & 24.2                          & 22.0                          \\
THID\cite{wang2022learning} & 15.5                           & 24.3                          & 23.0                          \\
SCL\cite{hou2022scl} & 19.1                            & 30.4                          & 28.1                          \\
DDS (Ours)  & \textbf{21.1}                    & \textbf{31.7}                  & \textbf{29.6}  
    \\
\hline 
\end{tabular}
}

\vspace{0.2cm}
\caption{ DDS's performance comparison in HICO-Det test set under RF (Rare-First) compositional setting. The best results are shown in \textbf{bold}.  }
\label{tab:map_hico}
\vspace{-0.2cm}
\end{table}

\begin{table}[t]
\centering
\adjustbox{width=\columnwidth}{

\begin{tabular}{lccc}
\hline
Method                                        & Unseen (mAP)  & Seen (mAP)    & Full (mAP)    \\ \hline
VRD\cite{lu2016visual}                       & -                                  & -                                  & 7.2                               \\
WSL\cite{peyre2017weakly}                    & -                                 & -                                 & 9.9                               \\
DUV\cite{peyre2019detecting} & -                                 & -                                 & 13.4                              \\
DDS (Ours)                           & \multicolumn{1}{c}{\textbf{16.3}} & \multicolumn{1}{c}{\textbf{27.4}} & \multicolumn{1}{c}{\textbf{17.9}} \\
\hline
\end{tabular}
}
\vspace{0.2cm}
\caption{ DDS's performance comparison in UnRel test set. The best results are shown in \textbf{bold}. Results not reported are marked with a  '-'.}
\vspace{-0.2cm}
\label{tab:map_unrel}
\end{table}

\subsection{Comparison with the SOTA models} \label{sec:sota}

In the AG\cite{ji2020action} dataset, we report DDS's performances in Table \ref{tab:ag_sota_comp} under the compositional setting. In this setting, there are $\sim 12\%$ less training data with $80$ unseen relationship triplets. We retrain STTRan\cite{cong2021spatial} in the mentioned setting for comparison. It is important to note here, that among the three recent DSG generation models\cite{ji2021detecting, cong2021spatial, li2022dynamic}, only STTran's code is publicly available, therefore limiting the capacity to evaluate other models. DDS outperforms STTran in all recall levels in both unseen and seen relationship triplet detection. Especially, for detecting unseen triplets DDS achieves $4-24$ times improvement over the SOTA model. It shows the generalization power of DDS.

Additionally, for a fair comparison with other models, we train DDS in the full training set of AG dataset and report performance in Table \ref{tab:map_AG}. With similar training data, DDS achieves SOTA performance on all metrics, especially at R@10,  thus requiring fewer predictions to capture the actions in the scene. Among the compared works, HORT \cite{ji2021detecting} has a mutlti-branch Transformer based architecture which is similar to DDS on a high level. However, the primary innovation of DDS lies in its ability to learn object and relationship features separately. 
Since HORT has not made their code publicly available, a direct comparison of performance in detecting unseen triplets is not feasible.\ER{However, considering the significant improvements demonstrated by our method, as shown in Table \ref{tab:map_AG} — an average recall enhancement of 7\% in the fully supervised settings — underscoring its efficacy. We anticipate that this performance gap would be even higher in the context of detecting unseen triplets. Similarly, most recent papers as \cite{Lin2024TD2NetTD},\cite{VsCGG}, and \cite{APT} are not designed with the end goal of predicting unseen triplets. Given that our method outperforms them by a significant margin using the full data of AG, we would expect a significant performance gap in the compositional setting as well, however, a comparison is again limited due to the lack of a publicly available implementation of their approach.}

In Table \ref{tab:map_hico} and Table \ref{tab:map_unrel}, DDS outperforms all existing methods in HICO-Det\cite{chao2018learning} and UnRel\cite{peyre2017weakly} datasets. In summary,  DDS outperforms existing works in both seen and unseen SSG generation (full category) in HICO-Det by $5\%$ and in UnRel by $33\%$. \ER{It is worth noting that different from the works in \cite{cao2023re,tip1,tip1,tip3}, we use the mentioned datasets to assess only our performance under the compositional setting to evaluate DDS' capability of predicting unseen triplets.}

\subsection{Ablation Studies}\label{sec:abl}
We perform ablations for different design choices of our network in this section. Other than the temporal decoder ablation, we report our performances in the HICO-Det\cite{chao2018learning} dataset. We utilize AG dataset for the temporal decoder ablation.
The complete tables with the results of each ablation study are presented in the supplemental material. 

\begin{figure*}[t]
\begin{center}
\includegraphics[width=\linewidth]{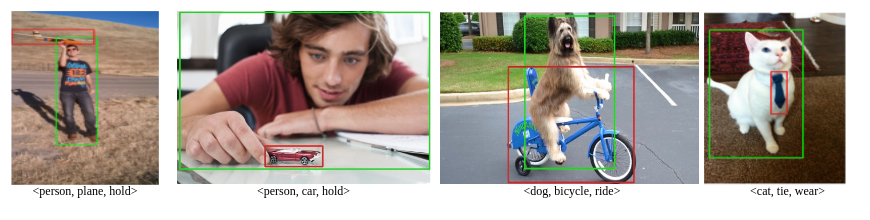}
\end{center}
 \vspace{-0.4cm}
\caption{Qualitative results of DDS for predicting unusual relationship triplets in UnRel\cite{peyre2017weakly} dataset. The subject bounding box is green and the object bounding box is red.}
  \vspace{-0.2cm}
\label{fig:qual1}
\end{figure*} 


\textbf{Multi-branch Design:} We first validate the decoupled multi-branch design and present the results in \ref{tab:ab_hico_parts}. The single-branch network, which uses a shared set of queries for both object and relationship detection, performs poorly, especially on unseen categories. Adding two spatio-temporal decoders improves performance, and further introducing two separate encoders and relation region prediction yields significant gains. This demonstrates the benefit of decoupled learning.

\begin{table}[h]
\centering
\small
\setlength{\tabcolsep}{4pt}
\adjustbox{width=\columnwidth}{

\begin{tabular}{lccccc}
\hline
Type                   & \begin{tabular}[c]{@{}c@{}}Relation \\ Region\end{tabular} & \begin{tabular}[c]{@{}c@{}}Separate \\ Encoders\end{tabular} & \begin{tabular}[c]{@{}c@{}}Separate \\ Decoders\end{tabular} & Unseen (mAP)  & Seen (mAP)    \\ \hline
Base network                 & \xmark                                                     & \xmark                                                      & \xmark                                                                      & 17.9          & 29.9          \\ \hline
\multirow{3}{*}{Multi branch} & \xmark                                                     & \xmark                                                      & \cmark                                                                      & 18.7          & 30.5          \\
                              & \xmark                                                     & \cmark                                                      & \cmark                                                                      & 19.7          & 31.6          \\
                              & \cmark                                                     & \cmark                                                      & \cmark                                                                      & \textbf{21.1} & \textbf{31.7} \\ \hline
\end{tabular}

}
\caption{ Impact of different components on our decoupled multi-branch design.}
\label{tab:ab_hico_parts}
\vspace{0.4cm}
\end{table}

\textbf{Relation Region Ground Truths:} During training, the relation branch requires ground truth relation regions. We experiment with two approaches: (1) Mixture, where the relation region is the intersection of subject and object boxes if IoU exceeds $\theta$, otherwise their union, and (2) Union box, where the union of subject and object boxes defines the relation region. We noticed that performance is correlated with the value of $\theta$ and the best value is the union box, which corresponds to $\theta = 1$, likely due to its inclusion of spatial information for non-contact relationships (e.g., subject looks at object).

\textbf{Share Queries:} We test two query-sharing strategies between the object and relation branches: (1) o to r (object to relation) and (2) r to o (relation to object). DDS performs best without shared queries, especially for unseen triplets, reinforcing the importance of decoupled queries. Sharing queries, particularly from relation to object, reduces performance, indicating that object queries generalize better.

\textbf{Spatial Decoders:} We evaluate different numbers of layers for the spatial decoders. Performance suffers with too few or too many layers, with the best results when the object decoder has more layers than the relation decoder, reflecting the need for more capacity when decoding two entities (subject and object).

\textbf{Temporal Decoders:} Temporal decoders play a crucial role in detecting relationship triplets. Without them, performance drops by about 3\% recall, confirming that temporal dependencies across frames are key for accurate triplet predictions.

\subsection{Qualitative Results}\label{sec:qual}

This section compares DDS's performance with our base network (details in section \ref{sec:abl}). This comparison is made using the UnRel\cite{peyre2017weakly} dataset.
Fig. \ref{fig:qual1} illustrates the samples where DDS is successful; however, our base network predicted bounding boxes do not match with ground truth. More results can be seen in the supplemental materials.

\begin{figure}[h]
\begin{center}
\includegraphics[width=1\linewidth]{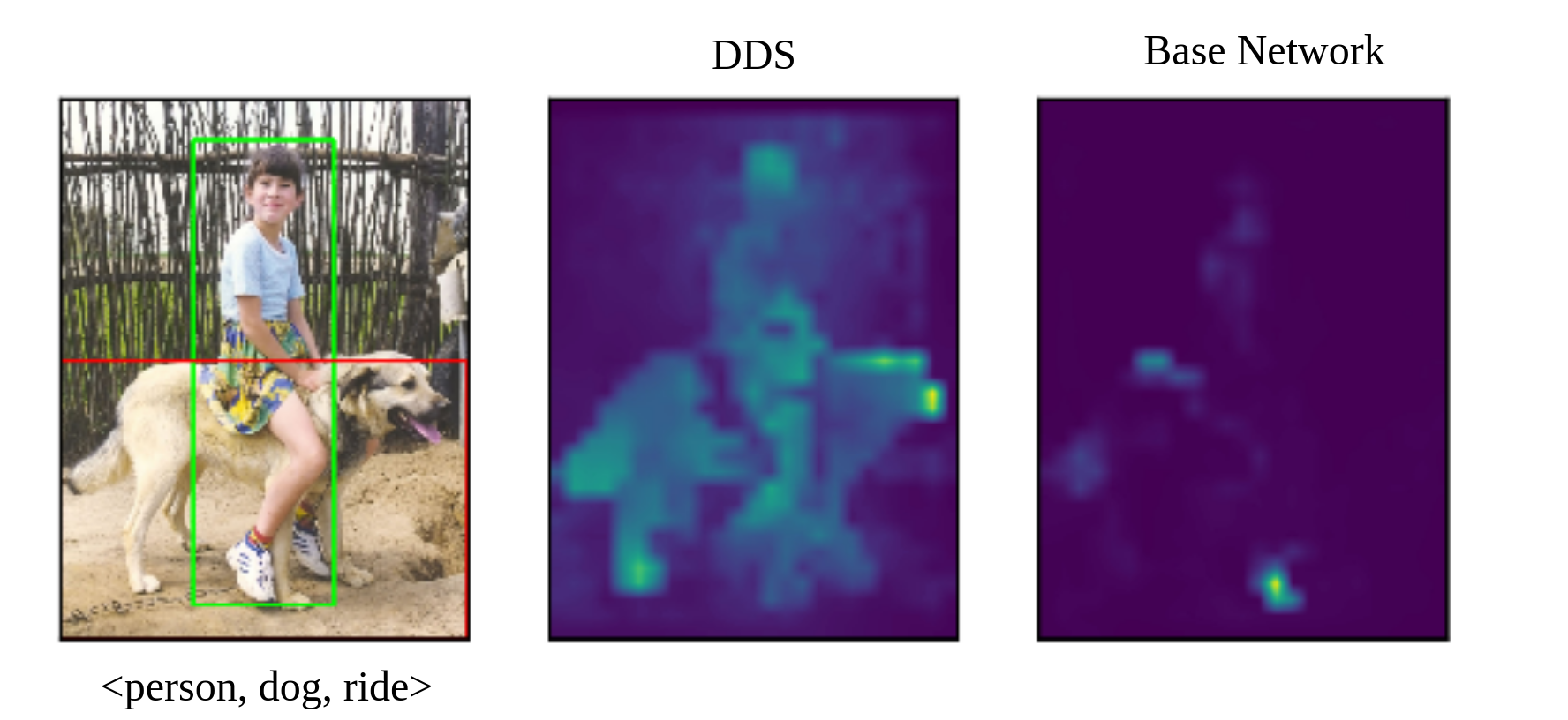}
\end{center}
 \vspace{-0.4cm}
\caption{Performance analysis of DDS over the base network. The attention maps are visualized from the last layer of the spatio-temporal decoder.}
  \vspace{-0.2cm}
\label{fig:qual2}
\end{figure}

We compare the attention maps from DDS and the base network to further analyze our improved performances. Fig. \ref{fig:qual2} presents attention maps for the samples where both DDS and the base network have correct predictions. The attention maps are of the queries that predict the marked subject and object bounding boxes from the last layer of the spatio-temporal decoders. We overlap attention maps from both our spatio-temporal decoders to get the final attention map. As can be seen, although both networks have correct predictions, DDS's attention maps cover the correct spatial region compared to spotty locations produced by the base network. 

\section{Conclusion}
This paper proposes DDS, a multi-branch decoupled network for DSG generation. it is comprised of two encoder-decoder based Transformer branches. This design enables independent learning of objects and relationships, thus improving the performance when detecting unseen relationship triplets. The effectiveness of DDS is demonstrated through extensive experiments where it achieves SOTA performance on three benchmark datasets. Moreover, the conducted ablation studies have provided the motivation and significance for different components of DDS. However, while successful compared to the existing works, the quantitative results show room for improvement in detecting unseen relationship triplets. Future research will focus on improving DDS for a better generalized DSG generation.
\section*{Acknowledgments}
This research is partially supported by the following grants: US Army Research Laboratory (ARL) under agreement number W911NF2020157; and by NSF award SI2-SSI \#1664172. The U.S. Government is authorized to reproduce and distribute reprints for Governmental purposes notwithstanding any copyright notation thereon. The views and conclusions contained herein are those of the authors and should not be interpreted as necessarily representing the official policies or endorsements, either expressed or implied, of US Army Research Laboratory (ARL) or the U.S. Government. 

{\small
\bibliographystyle{ieee_fullname}
\bibliography{egbib}
}

\end{document}